\documentclass[11pt]{article}

\usepackage[margin=1in]{geometry}
\usepackage{amsmath, amssymb, amsthm}
\usepackage{bm}
\usepackage{graphicx}
\usepackage{booktabs}
\usepackage{hyperref}
\usepackage{enumitem}
\usepackage{natbib}
\usepackage{xcolor}
\usepackage{titlesec}
\usepackage{setspace}
\usepackage{caption}
\usepackage{subcaption}

\hypersetup{
  colorlinks=true,
  linkcolor=blue!60!black,
  citecolor=blue!60!black,
  urlcolor=blue!60!black
}

\titleformat{\section}{\large\bfseries}{\thesection}{0.6em}{}
\titleformat{\subsection}{\normalsize\bfseries}{\thesubsection}{0.6em}{}

\newcommand{\N}{\mathcal{N}}
\newcommand{\dt}{\Delta t}

\title{\bfseries DTVEM-RE: A Hierarchical Random-Effects Extension of the Differential Time-Varying Effect Model for Person-Specific Multi-Lag Estimation in Intensive Longitudinal Data}

\author{Amartya Bhattacharya\\[0.3em]
\small Geisel School of Medicine, Dartmouth College}

\date{\today}

\begin{document}

\maketitle

\begin{abstract}
\noindent
The Differential Time-Varying Effect Model (DTVEM) introduced by Jacobson, Chow, and Newman (2019) has become a widely-used tool for identifying optimal lag structures in intensive longitudinal data, pairing a generalized additive mixed model (GAMM) exploratory stage with a state-space vector autoregression (VAR) confirmatory stage. However, DTVEM as published assumes a single group-level lag structure shared across individuals, a limitation that the original authors named as future work and one that conflicts with the idiographic premise of much modern psychopathology research. We present DTVEM-RE, a hierarchical random-effects extension of DTVEM that estimates person-specific multi-lag coefficients with shrinkage toward group-level means via Hamiltonian Monte Carlo in Stan. We report three sets of results. First, a simulation study across four heterogeneity levels confirms clean recovery of the between-person variance parameter ($\tau_a$) with absolute bias below 0.01 and credible interval coverage of 90 to 93 percent at sample sizes typical of EMA studies. Second, a three-item empirical demonstration on the Fisher et al. (2017) ecological momentary assessment dataset ($N = 40$ outpatients with generalized anxiety and major depressive disorder) shows that person-specific lag-1 autoregressive effects vary by an order of magnitude across affect items, that hierarchical Bayesian and independent GAMM estimates of person-specific coefficients agree closely ($r = 0.87$ to $0.92$), and that DTVEM-RE achieves the best one-step-ahead predictive log-likelihood and root-mean-square error among four hierarchical and non-hierarchical baselines, albeit by modest margins. Third, a multi-lag extension shows that all nine $\tau_k$ estimates across three items and three lags have 90 percent credible intervals excluding zero, with the lag at which heterogeneity is largest differing across items. This multi-lag person-specific heterogeneity is outside the modeling scope of existing hierarchical VAR methods that estimate random effects only on lag-1 coefficients, such as mlVAR. We conclude that DTVEM-RE provides, to our knowledge, the first principled idiographic implementation of DTVEM-style lag detection while retaining standard DTVEM as a fixed-effects special case.
\end{abstract}

\vspace{0.5em}

\noindent\textbf{Keywords:} DTVEM; hierarchical state-space model; ecological momentary assessment; idiographic psychopathology; random effects; intensive longitudinal data; mood and anxiety dynamics

\vspace{1em}

\section{Introduction}

\subsection{Idiographic dynamics in clinical psychology}

The recognition that individuals differ meaningfully in their psychological dynamics has reshaped quantitative clinical research over the past two decades. \citet{molenaar2004manifesto} formalized this concern in a non-ergodicity argument, demonstrating that conclusions drawn from between-person variation generally do not transfer to within-person dynamics. The implication is methodologically consequential: standard cross-sectional and group-based longitudinal models can misrepresent the very phenomena they are designed to study when the question concerns processes unfolding within an individual.

Empirical evidence for this position has accumulated rapidly. \citet{fisher2017exploring} provided a particularly influential demonstration in a sample of 40 outpatients with generalized anxiety disorder (GAD), major depressive disorder (MDD), or both, showing that person-specific symptom networks differed substantially from group-aggregated patterns. \citet{fisher2018lack} formalized this into a generalizability critique, arguing that lack of group-to-individual transferability is a structural threat to human subjects research that cannot be solved through larger samples alone. Subsequent work has built personalized treatment-targeting algorithms on top of idiographic dynamic models \citep{fernandez2017dynamic, fisher2019open}, contributing to a broader shift toward person-centered clinical methodology.

\subsection{The DTVEM framework and its group-level limitation}

A core question in this research program concerns the temporal architecture of symptom dynamics: at what time interval does one symptom or mood predict another? The Differential Time-Varying Effect Model (DTVEM) introduced by \citet{jacobson2019differential} addresses this question through an elegant two-stage hybrid procedure.

In Stage 1, an exploratory generalized additive mixed model (GAMM) smooths the autoregressive coefficient across lag distance $\dt$, producing a continuous lag-effect curve $\hat{f}(\dt)$. Peaks and valleys in this curve, where the 95 percent confidence band excludes zero, identify candidate lags at which dependencies appear statistically reliable. In Stage 2, a vector autoregression (VAR) model is specified using only those candidate lags as predictors, then estimated as a state-space model in OpenMx. The result is a confirmatory model with only the lags that Stage 1 has identified as supported by the data.

DTVEM has been widely adopted across psychopathology, sleep, personality, and digital biomarker research. Its appeal lies in combining nonparametric flexibility (Stage 1 makes few assumptions about lag shape) with parsimonious confirmatory estimation (Stage 2 fits only the structure the data supports).

However, DTVEM as published assumes that the lag structure $\hat{f}(\dt)$ is shared across all individuals in the dataset. The exploratory smoother is fit to pooled data, and the confirmatory VAR estimates a single set of coefficients applied uniformly. The original authors explicitly acknowledged this limitation in their concluding remarks:

\begin{quote}
``The present simulation studies did not consider person-specific differences in dynamics or lag structures. Given this, future extensions of the DTVEM model may be able to include random effects within the GAMM framework to model person-specific differences'' \citep[p.~311]{jacobson2019differential}.
\end{quote}

This limitation is more than technical. The premise of idiographic clinical research is that individuals differ in their dynamics. Applying a method that assumes shared lag structure to a research question fundamentally about heterogeneity is internally inconsistent: it discards the very variation the research is designed to study. The methodological gap is precisely the one named by the original authors as a priority for future work.

\subsection{The present work}

We present DTVEM-RE (DTVEM with Random Effects), a hierarchical extension that estimates person-specific lag profiles with shrinkage toward a group-level mean. The exploratory stage uses factor-smooth GAMMs to recover individual lag curves with automatic per-person regularization. The confirmatory stage embeds those lags in a hierarchical Bayesian state-space VAR estimated via Hamiltonian Monte Carlo in Stan.

The contributions are threefold. First, we specify the DTVEM-RE model and provide a Stan implementation, with full code released for replication. Second, we validate parameter recovery through a controlled simulation study, showing that the between-person variance parameter $\tau_a$ is recovered without meaningful bias and with near-nominal credible interval coverage at the sample sizes typical of EMA studies. Third, we apply DTVEM-RE to the Fisher et al. (2017) ecological momentary assessment dataset and demonstrate (a) substantial between-person heterogeneity in lag-1 autoregressive effects across three affect items, (b) modest but consistent predictive improvement over alternative methods in held-out evaluation, and (c) statistically robust between-person heterogeneity at lags 1, 2, and 3 simultaneously in a multi-lag extension, with the lag at which heterogeneity is largest differing across items. This last finding is outside the modeling scope of existing hierarchical VAR methods that place random effects only on lag-1 coefficients.

The paper proceeds as follows. Section 2 reviews related methods and positions DTVEM-RE within the existing landscape. Section 3 specifies the data, the model, the priors, the estimation procedure, and the simulation and held-out prediction protocols. Section 4 reports results in four parts: parameter recovery in simulation, the Stage 1 GAMM exploration on Fisher's data, the lag-1 Stage 2 fits with cross-method validation and held-out prediction, and the multi-lag extension. Section 5 discusses implications, limitations, and directions for future work.

\section{Related Methods}

DTVEM-RE sits at the intersection of two methodological literatures: hierarchical extensions of VAR models for intensive longitudinal data, and lag-exploration methods that go beyond fixed lag-1 specifications. We briefly review each.

\subsection{Hierarchical lag-1 methods}

The dominant hierarchical extension of VAR models for idiographic psychopathology research is multilevel VAR, or mlVAR \citep{bringmann2013network, epskamp2018personalized}. The framework originated with \citet{bringmann2013network}, which introduced a multilevel VAR with random effects on lag-1 autoregressive and cross-lagged coefficients for ESM data, and was extended by \citet{epskamp2018personalized} to combine temporal and contemporaneous network estimation. mlVAR is widely used and has informed many of the substantive findings in the network psychopathology literature. As implemented, however, it estimates random effects only on lag-1 coefficients; it is not designed to discover whether between-person heterogeneity differs in magnitude across lag distances.

Group Iterative Multiple Model Estimation (GIMME) and its latent-variable extension LV-GIMME \citep{gates2020latent} take a different approach to heterogeneity, performing a data-driven search for person-specific structural paths. Lagged paths can be included, and GIMME does not constrain itself to lag-1 paths in principle. However, GIMME is a model-search method rather than a hierarchical-pooling method; it does not place a population distribution on person-specific lag coefficients and does not provide the shrinkage that hierarchical models do at moderate per-person sample sizes.

Multi-VAR \citep{fisher2022multivar} estimates VAR models for multiple subjects simultaneously using adaptive-LASSO penalization, allowing for shared and idiosyncratic transition-matrix elements across persons. The framework supports multiple lags, but it does not place a hierarchical population distribution on person-specific lag coefficients with explicit shrinkage. The mechanism for handling heterogeneity is penalized estimation rather than partial pooling.

\subsection{Multi-lag exploration methods}

DTVEM \citep{jacobson2019differential} is the leading method for nonparametric lag exploration in intensive longitudinal data. Its GAMM-based Stage 1 makes no parametric assumption about the shape of the lag-effect kernel, allowing detection of complex structures including delayed effects, secondary peaks, and oscillatory patterns. However, DTVEM operates at the group level only.

Continuous-time alternatives, including hierarchical ctsem \citep{driver2018ctsem}, model the underlying process in continuous time and naturally handle irregular spacing. These methods impose an exponential decay shape by construction, sacrificing the nonparametric flexibility of DTVEM in exchange for principled handling of unequal time intervals. Hierarchical ctsem supports random effects across all model parameters, including the drift matrix that governs temporal dependence, but the lag-effect shape is fixed by the continuous-time formulation rather than estimated nonparametrically.

\subsection{The gap DTVEM-RE fills}

To our knowledge, no published method combines nonparametric multi-lag exploration with hierarchical pooling on person-specific lag coefficients at each detected lag. The closest existing approaches each cover one half of this combination:

\begin{itemize}[leftmargin=*,nosep]
    \item Standard DTVEM: multi-lag exploration, group-level only.
    \item mlVAR: hierarchical pooling, but on lag-1 random effects only.
    \item LV-GIMME and multi-VAR: support multiple lags and person-specific structure, but handle heterogeneity via model search or penalization rather than hierarchical pooling.
    \item Hierarchical ctsem: hierarchical and continuous-time, but with exponential decay imposed by the continuous-time formulation.
\end{itemize}

\section{Methods}

This section unifies the data, model specification, estimation procedure, and simulation design. The presentation order is substantive material first (what we analyzed) followed by methodological material in service of it. Stage 1 of DTVEM-RE is a hierarchical GAMM; Stage 2 is a hierarchical Bayesian state-space VAR. When the between-person variance components in Stage 2 are constrained to zero, DTVEM-RE reduces to standard DTVEM.

\subsection{Data}

We use the publicly available Fisher et al.\ (2017) ecological momentary assessment (EMA) dataset, hosted on the Open Science Framework at \url{https://osf.io/zefbc/}. The sample comprises 40 outpatients meeting DSM criteria for generalized anxiety disorder (GAD), major depressive disorder (MDD), or both. Each participant completed smartphone-delivered EMA prompts four times daily at pseudo-random intervals within waking hours, for approximately thirty days. Each prompt collected $0$ to $100$ visual-analog ratings on twenty-eight affect and anxiety items.

After dropping missed prompts, the analytic dataset comprised 4,463 completed beeps across the forty participants, with a median of 113 beeps per person (range 90 to 151). Two participants had within-protocol compliance breaks exceeding seventy-two hours; we segmented these into separate sessions and used only the first session per participant throughout, in order to maintain comparable analytic windows across the sample. Median spacing between consecutive completed beeps was approximately 4.3 hours during the day, with overnight gaps of 10 to 12 hours.

We focus on three items chosen to span affect domains: {\it down} (negative affect, low arousal; representative of the depression-core cluster), {\it worried} (negative affect, high arousal; representative of the anxiety-core cluster), and {\it energetic} (positive activation). Demonstrating the method on three items rather than one establishes that results are not item-specific. We report all three items in full at every stage of analysis. We did not pre-screen items for the property under study (between-person heterogeneity in lag coefficients); these three were selected on substantive grounds before estimation.

\begin{figure}[ht]
\centering
\includegraphics[width=\textwidth]{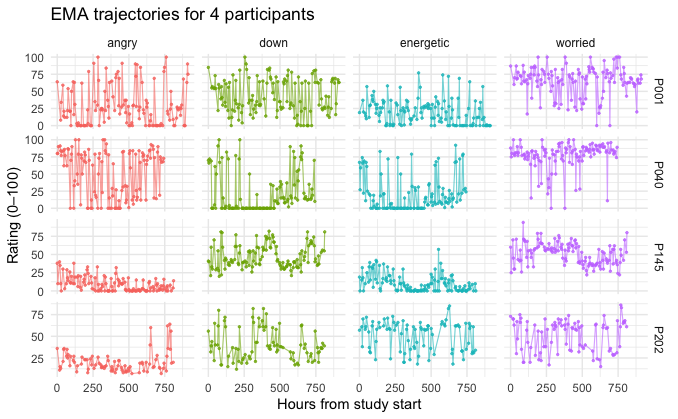}
\caption{EMA item trajectories for four representative participants (P001, P040, P145, P202) across the full study period. Three of the four items shown (\textit{down}, \textit{worried}, \textit{energetic}) are analyzed throughout this paper; \textit{angry} is included for additional illustration of the qualitative range of dynamics in this dataset. The horizontal axis displays hours from study start; the analyses reported below use sequential beep number as the lag index (see Section 3.2). The four participants were selected to span the range of dynamic profiles in the sample, from high-amplitude and volatile (P001) to low-amplitude and tightly bounded (P145). Participant identifiers are the original IDs from the Fisher et al.\ (2017) dataset.}
\label{fig:trajectories}
\end{figure}

Figure~\ref{fig:trajectories} shows EMA item trajectories for four representative participants, illustrating the qualitative range of dynamics this paper aims to characterize.

\subsection{Notation and preprocessing}

Let $y_{i,t}$ denote the rating from participant $i$ ($i = 1, \ldots, N = 40$) at beep occasion $t$ ($t = 1, \ldots, T_i$). Within each participant, ratings were z-scored using the participant-specific mean and standard deviation. This within-person standardization removes individual differences in mean level and rating-scale use, allowing the analysis to focus on within-person fluctuation. Standardization additionally allows direct interpretation of AR coefficients across items measured on a common scale.

Lags are constructed using sequential beep number rather than wall-clock time. That is, $y_{i,t-1}$ denotes participant $i$'s rating at the beep occasion immediately preceding $t$, irrespective of whether that interval was approximately three hours (within-day) or twelve hours (overnight). This approximation treats beeps as approximately equally spaced, which holds within waking hours but not across the overnight gap. We adopt this convention because it matches the practice used by \citet{jacobson2019differential} and the broader EMA modeling literature. The implications are revisited in Section 5. Rows for which any of the requisite lagged values is missing are dropped from the analytic sample. For the lag-1 model this excludes the first within-session observation per participant and any row following a within-session gap; for the multi-lag model fit at lags 1 to 3 the excluded set is larger because $y_{i,t-1}$, $y_{i,t-2}$, and $y_{i,t-3}$ must all be observed.

\subsection{Stage 1: Hierarchical exploratory GAMM}

Stage 1 uses a generalized additive mixed model (GAMM) to estimate the lag-effect curve as a function of lag distance $\dt$. Standard DTVEM Stage 1 fits a single population-level smooth $f(\dt)$ to pooled data. We extend it to allow person-specific deviations from this population smooth:
\begin{equation}
y_{i,t} = \big[ f(\dt) + b_i(\dt) \big] \cdot y_{i, t-\dt} + \varepsilon_{i,t}, \qquad \varepsilon_{i,t} \sim \N(0, \sigma^2)
\label{eq:stage1}
\end{equation}
Here $f(\dt)$ is the population-level smooth lag-effect curve and $b_i(\dt)$ is a person-specific deviation from it. Both are represented via thin-plate regression splines. Implementation uses the factor-smooth basis ({\tt bs = "fs"}) in {\tt mgcv::bam} \citep{wood2017gam}, which couples all participant-specific smooths through a single shared smoothing parameter, providing automatic per-person shrinkage toward the population smooth: heavy shrinkage where individual data is sparse, less shrinkage where individual data is informative enough to support a distinct shape.

Stage 1 plays two roles in DTVEM-RE. First, it provides an exploratory view of which lags carry substantial dependence at the population level and how person-specific lag profiles deviate from that average. Second, it provides an independent estimate of person-specific lag-1 coefficients that serves as a cross-method check on Stage 2 posterior estimates.

We do not use Stage 1 to perform automated lag-set selection in the present paper. Stage 2 is fit with an analyst-specified lag set $\mathcal{K}$, with $\mathcal{K} = \{1\}$ and $\mathcal{K} = \{1, 2, 3\}$ both reported below. Automating the handoff from Stage 1 candidate lags to Stage 2 lag set is straightforward in principle and follows the design of standard DTVEM, but the additional methodological choices it raises (significance thresholding, multiple-lag selection rules) are kept out of scope here to focus this paper on the hierarchical pooling step.

\subsection{Stage 2 (lag-1 case): hierarchical Bayesian state-space VAR}

In the present paper we report results for the univariate special case in which a single item is modeled at a time. The framework extends directly to multivariate VAR by replacing the scalar coefficient $a_i$ with a transition matrix $\mathbf{A}_i$ at each lag, with hierarchical priors on the matrix entries. We restrict attention here to the univariate case to focus the methodological development on the hierarchical pooling step; the multivariate extension is a natural follow-up.

For the single-lag case, Stage 2 places a hierarchical normal distribution over person-specific autoregressive coefficients $a_i$ and a hierarchical log-normal distribution over person-specific residual standard deviations $\sigma_i$:
\begin{align}
y_{i,t} &\sim \N\big(a_i \cdot y_{i,t-1},\ \sigma_i^2\big) \label{eq:lik}\\
a_i &= \mu_a + \tau_a \cdot a^{\text{raw}}_i, \quad a^{\text{raw}}_i \sim \N(0, 1) \label{eq:nca}\\
\sigma_i &= \sigma_{\text{mean}} \cdot \exp(\sigma_{\text{sd}} \cdot \sigma^{\text{raw}}_i), \quad \sigma^{\text{raw}}_i \sim \N(0, 1) \label{eq:ncs}
\end{align}
Equations~\eqref{eq:nca} and~\eqref{eq:ncs} implement non-centered parameterizations of the person-specific coefficient and the person-specific residual SD. Non-centered parameterizations reparameterize the relationship between a person-level parameter and its population-level scale to a unit-scale auxiliary variable, breaking the funnel-shaped posterior pathology that hierarchical models exhibit when the scale parameter is small relative to the per-person sample size \citep{betancourt2015hamiltonian}. In an initial development version of the model, the centered parameterization for $\sigma_i$ produced 127 divergent transitions across four chains on the {\it energetic} item; the non-centered log-normal reparameterization eliminated all divergences without altering parameter estimates beyond Monte Carlo error.

The hyperparameters are $\mu_a$, the population-level mean of the AR(1) coefficient; $\tau_a$, the between-person standard deviation of the AR(1) coefficient and the parameter of primary substantive interest; $\sigma_{\text{mean}}$, the population-level residual standard deviation; and $\sigma_{\text{sd}}$, the between-person variation in residual SDs on the log scale. The substantive yield of DTVEM-RE relative to standard DTVEM is captured by $\tau_a$: when $\tau_a = 0$, the model reduces to a fixed-effects AR(1) shared across persons.

\subsection{Stage 2 (multi-lag case)}

The single-lag model extends to $K$ lags by giving each person a vector of $K$ coefficients with its own hyperparameters:
\begin{align}
y_{i,t} &\sim \N\Bigg(\sum_{k=1}^{K} a_{i,k} \cdot y_{i, t-k},\ \sigma_i^2\Bigg) \\
a_{i,k} &= \mu_k + \tau_k \cdot a^{\text{raw}}_{i,k}, \quad a^{\text{raw}}_{i,k} \sim \N(0, 1), \quad k = 1, \ldots, K
\end{align}
with $\sigma_i$ retaining the log-normal hierarchical specification above. Each lag receives its own population mean $\mu_k$ and its own between-person standard deviation $\tau_k$. This is the design choice that distinguishes DTVEM-RE from a multi-lag extension constraining heterogeneity to be uniform across lags. The separate-$\tau_k$ specification allows the {\it shape} of between-person heterogeneity across lags to differ from the {\it shape} of population means across lags, and is what makes the empirical result in Section 4.5 possible.

The raw deviations $a^{\text{raw}}_{i,k}$ are sampled independently across lags in the present specification; we do not place a prior on cross-lag covariance within $\mathbf{a}_i = (a_{i,1}, \ldots, a_{i,K})^\top$. A more flexible specification would estimate a $K \times K$ covariance matrix over the person-level coefficient vector and is a natural extension. We report results from the independent specification throughout and examine empirical cross-lag correlations among posterior means in Section 4.5.

\subsection{Priors and rationale}

The model places weakly informative priors on the hyperparameters:
\begin{align}
\mu_a &\sim \N(0,\ 0.5), \qquad \tau_a \sim \text{Half-}\N(0,\ 0.3) \\
\mu_k &\sim \N(0,\ 0.5), \qquad \tau_k \sim \text{Half-}\N(0,\ 0.3) \quad (k = 1, \ldots, K) \\
\sigma_{\text{mean}} &\sim \N(0.9,\ 0.3), \qquad \sigma_{\text{sd}} \sim \text{Half-}\N(0,\ 0.3)
\end{align}

The $\N(0, 0.5)$ prior on the population means $\mu_a$ and $\mu_k$ is centered at zero (no temporal dependence) with a standard deviation of 0.5 on the within-person z-scored response scale. This places approximately 95\% prior mass between $-1$ and $+1$, comfortably covering the stationary range $(-1, 1)$ for an AR(1) coefficient on a standardized series while pushing posterior mass away from the unit-root boundary. The Half-$\N(0, 0.3)$ prior on the between-person SDs $\tau_a$ and $\tau_k$ places approximately 95\% prior mass below $0.6$, an upper bound chosen to be substantially larger than the magnitudes of between-person variation reported in the broader EMA affect-dynamics literature (typically well below $0.3$ on standardized AR coefficients) while concentrating prior density near zero to express skepticism toward implausibly large heterogeneity. We note that this prior was chosen before the empirical fits in Section 4.3 and reflects expectations from prior work rather than the observed values in the present sample. The $\N(0.9, 0.3)$ prior on $\sigma_{\text{mean}}$ reflects the expectation that residual SDs on a unit-variance standardized series will be near 1 if the AR(1) coefficient is small and somewhat less if it is large; the prior accommodates either case.

A note on identifiability is warranted. With $N = 40$ participants, estimating a hierarchical between-person SD $\tau_a$ requires care: at small $N$, half-normal priors on hierarchical scale parameters can produce posteriors that are partially prior-driven, especially when the empirical variance is small relative to the prior scale. The simulation study reported in Section 4.1 addresses this directly by examining $\tau_a$ recovery under known generating values across the empirically relevant range. The simulation results show that $\tau_a$ is recovered without meaningful bias and with approximately nominal credible interval coverage at $N = 40$, $T = 130$.

\subsection{Estimation}

Models are estimated via Hamiltonian Monte Carlo in Stan through the {\tt cmdstanr} interface. Default settings use 4 chains of 1000 warmup plus 1000 sampling iterations, with {\tt adapt\_delta = 0.95}. Convergence is assessed via $\hat{R} \leq 1.01$, divergent transition count, and energy Bayesian fraction of missing information (E-BFMI). Person-specific posterior summaries are extracted as posterior means with 90 percent credible intervals throughout. Lag-1 model fits complete in approximately 15 seconds per item on a standard desktop; multi-lag fits at $K = 3$ complete in approximately 2 minutes per item.

\subsection{Simulation study design}

We assess parameter recovery for DTVEM-RE under controlled conditions matched to the structure of the empirical data. Synthetic datasets were generated with $N = 40$ persons, $T = 130$ observations each, true population mean $\mu_a = 0.40$, and four between-person SD levels $\tau_a \in \{0, 0.10, 0.20, 0.30\}$. The range spans from the null case (no heterogeneity) through the moderate heterogeneity level observed empirically in Fisher's data ($\tau_a \approx 0.16$) up to a high-heterogeneity stress test.

For each condition, 30 independent replicates were generated. Within each replicate, person-specific true coefficients were sampled from $\N(\mu_a, \tau_a^2)$, constrained to the stationary range $[-0.95, 0.95]$. Person-specific residual standard deviations were sampled from a log-normal distribution with mean 0.9 and small variation, matching the structure assumed by the Stan model. AR(1) data was then generated for 130 time points per person, and the simulated series were demeaned within person (without rescaling) before fitting. The same Stan model used on Fisher's data (non-centered parameterizations on both $a_i$ and $\sigma_i$) was fit to each synthetic dataset.

Coverage is computed as the proportion of 30 replicates per condition in which the true generating value falls within the 90\% posterior credible interval. At the null condition $\tau_a = 0$, the half-normal prior excludes negative values by construction; coverage is therefore undefined for $\tau_a$ in this row and we report it as N/A.

\subsection{Held-out prediction protocol}

To assess whether DTVEM-RE's parameter estimates translate into improved predictive performance, we conducted a held-out prediction comparison on the {\it down} item. For each participant, the last 12 beeps (approximately three days of data) were held out as a test set; the remaining beeps formed the training set. Four methods were fit to the training data and used to produce one-step-ahead predictions for the held-out beeps:
\begin{enumerate}[leftmargin=*,nosep]
    \item {\bf Group-only DTVEM}: pooled OLS estimating a single autoregressive coefficient shared across persons.
    \item {\bf Naive per-person DTVEM}: separate OLS estimation per participant, no pooling.
    \item {\bf Hierarchical simple} (mlVAR-style baseline): hierarchical autoregressive model with random lag-1 coefficient and a single shared residual SD.
    \item {\bf DTVEM-RE (full)}: the lag-1 specification given in Section 3.4, with hierarchical lag coefficient and hierarchical person-specific residual SD.
\end{enumerate}
For each method, predictive performance is summarized by mean and total held-out log-likelihood across the 480 held-out observations and root-mean-square error.

\section{Results}

We report results in four parts. Section 4.1 reports the simulation study validating parameter recovery. Section 4.2 reports the Stage 1 GAMM exploration. Sections 4.3 and 4.4 report the lag-1 Stage 2 fits and the held-out prediction comparison against alternative methods. Section 4.5 reports the multi-lag extension.

\subsection{Simulation study: parameter recovery}

Table~\ref{tab:sim-results} summarizes parameter recovery across the 120 simulation fits (4 conditions $\times$ 30 replicates each).

\begin{table}[ht]
\centering
\small
\begin{tabular}{lrrrrrr}
\toprule
\textbf{True $\tau_a$} & $\hat\mu_a$ \textbf{(mean)} & \textbf{$\mu_a$ bias} & \textbf{$\mu_a$ cov.} & $\hat\tau_a$ \textbf{(mean)} & \textbf{$\tau_a$ bias} & \textbf{$\tau_a$ cov.} \\
\midrule
0.00 & 0.383 & $-$0.017 & 0.73 & 0.030 & $+$0.030 & N/A \\
0.10 & 0.387 & $-$0.013 & 0.83 & 0.096 & $-$0.004 & 0.90 \\
0.20 & 0.383 & $-$0.017 & 0.73 & 0.206 & $+$0.006 & 0.93 \\
0.30 & 0.364 & $-$0.036 & 0.87 & 0.291 & $-$0.009 & 0.93 \\
\bottomrule
\end{tabular}
\caption{Simulation recovery results across 4 conditions $\times$ 30 replicates. The between-person SD ($\tau_a$) is recovered with negligible bias and approximately nominal 90 percent coverage. The population mean ($\mu_a$) shows a small consistent downward bias of approximately 0.02.}
\label{tab:sim-results}
\end{table}

\begin{figure}[ht]
\centering
\begin{subfigure}{0.32\textwidth}
\centering
\includegraphics[width=\linewidth]{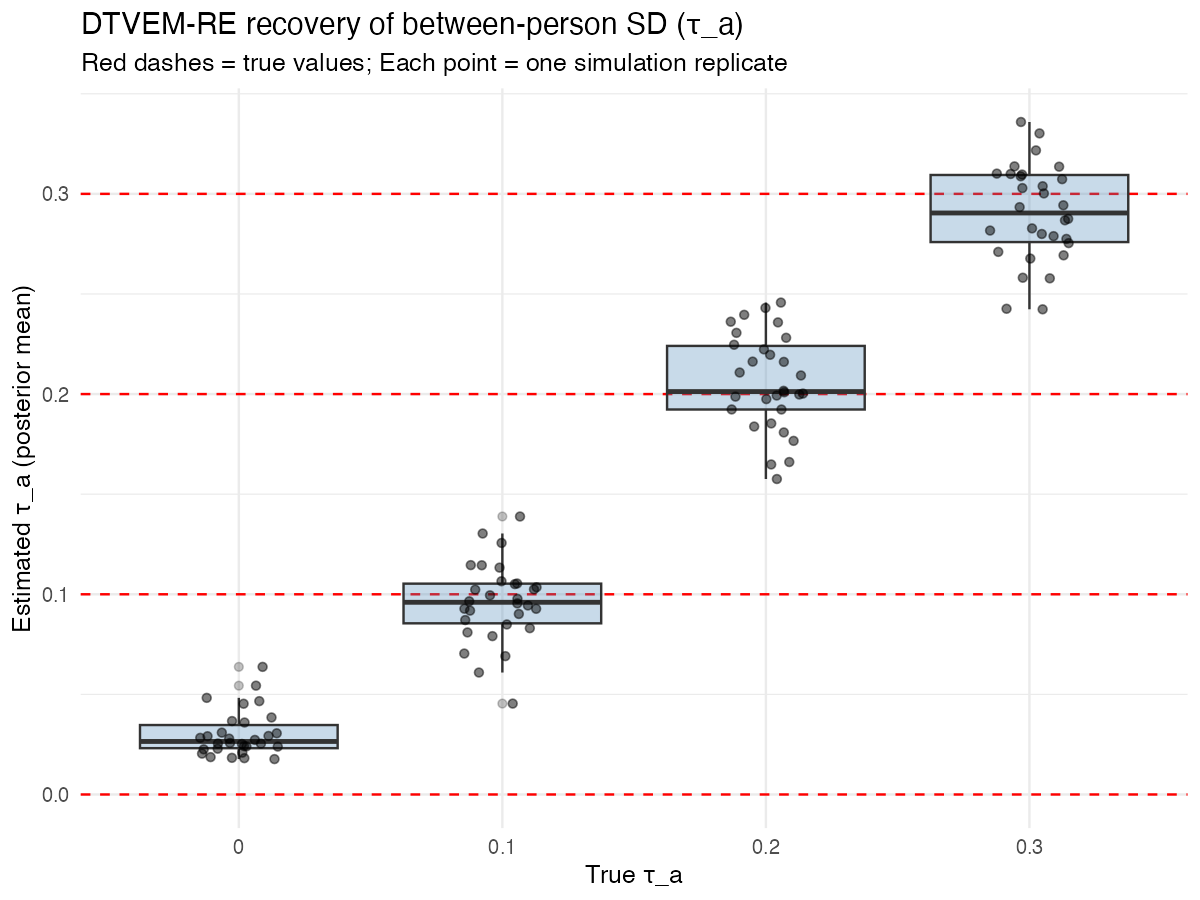}
\caption{Recovery of between-person SD ($\tau_a$). Each point is one of 30 simulation replicates; red dashes mark the true generating values.}
\label{fig:sim-tau}
\end{subfigure}
\hfill
\begin{subfigure}{0.32\textwidth}
\centering
\includegraphics[width=\linewidth]{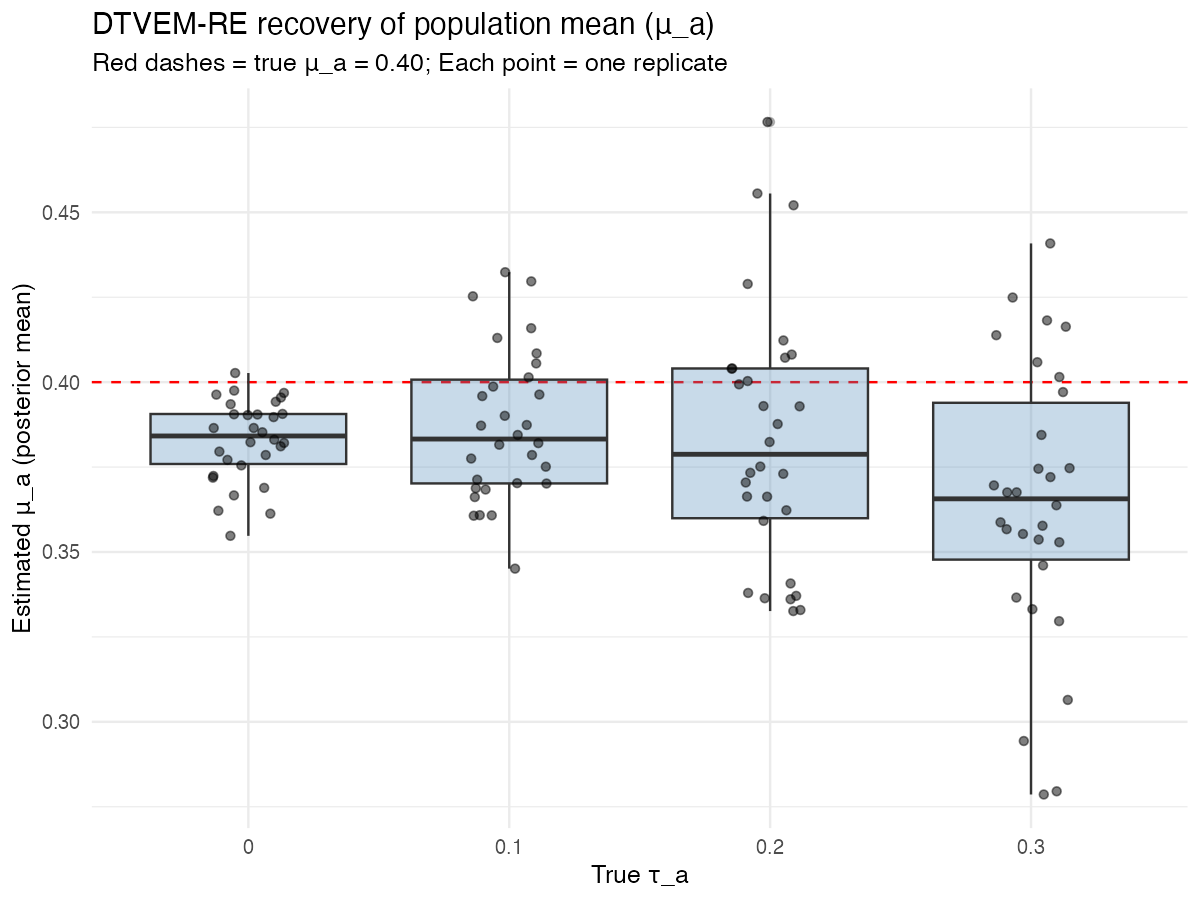}
\caption{Recovery of population mean ($\mu_a$). True $\mu_a = 0.40$ (red dashes) across all conditions.}
\label{fig:sim-mu}
\end{subfigure}
\hfill
\begin{subfigure}{0.32\textwidth}
\centering
\includegraphics[width=\linewidth]{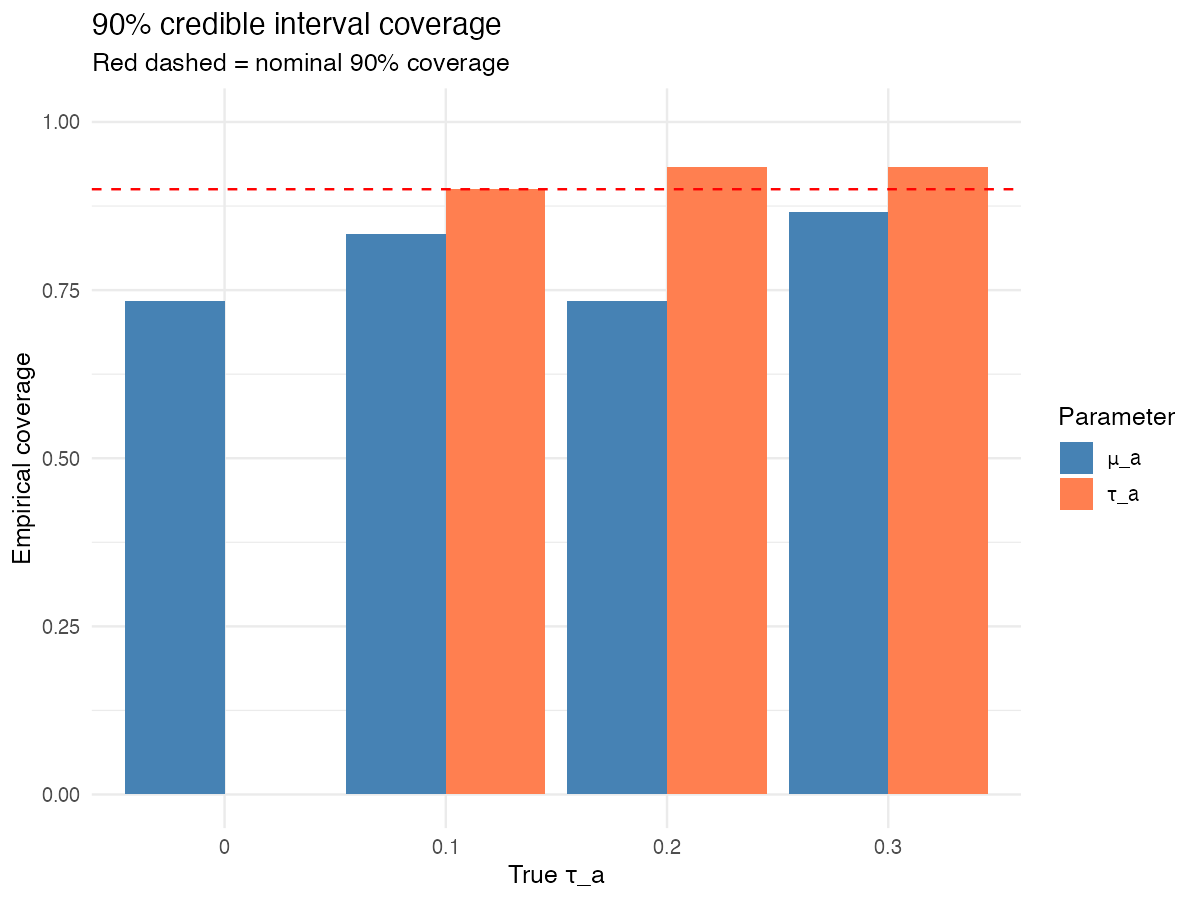}
\caption{Empirical 90\% credible interval coverage by parameter and condition; red dashed line marks nominal 90\%.}
\label{fig:sim-coverage}
\end{subfigure}
\caption{Parameter recovery from 120 simulation fits (4 conditions $\times$ 30 replicates), with $N = 40$ persons and $T = 130$ observations per person. Panel (a): $\tau_a$ posterior means recover the true generating values with negligible bias across non-null conditions. Panel (b): $\mu_a$ posterior means show a small consistent downward bias of approximately $0.02$ relative to the true $\mu_a = 0.40$. Panel (c): credible interval coverage is approximately at nominal 90\% for $\tau_a$ (orange) across non-null conditions, while $\mu_a$ coverage (blue) falls short of nominal due to the downward bias in panel (b). Coverage is undefined for $\tau_a$ at the null condition $\tau_a = 0$ because the half-normal prior excludes negative values.}
\label{fig:sim-recovery}
\end{figure}

Three findings emerge, visualized in Figure~\ref{fig:sim-recovery}. First, recovery of $\tau_a$, the parameter of primary methodological interest, is essentially unbiased. Absolute bias is at most 0.009 across all conditions with non-zero true heterogeneity. The half-normal prior produces a small positive estimate of 0.030 at the null condition by construction, since posterior mass concentrates near but cannot extend below zero. Second, $\tau_a$ credible interval coverage is 90 to 93 percent across non-null conditions, closely matching the nominal 90 percent target. This is the result that licenses interpreting the empirical $\tau_a$ estimates in Section 4.3 as reliable indicators of between-person heterogeneity rather than artifacts of the half-normal prior.

Third, $\mu_a$ shows a small consistent downward bias of approximately $0.02$, with empirical coverage 73 to 87 percent. The bias matches the leading-order finite-sample prediction $-(1 + 3a)/T \approx -0.017$ at $a = 0.4, T = 130$ \citep{marriott1954bias, kendall1954bias} closely at low-to-moderate heterogeneity conditions, with a slightly larger empirical bias ($-0.036$) at the high-heterogeneity stress condition $\tau_a = 0.30$. The bias affects all persons uniformly within a sample and would diminish as $T$ grows. Since DTVEM-RE's methodological claims center on between-person variation rather than the absolute level of the population mean, this bias does not threaten the substantive interpretation.

MCMC diagnostics were uniformly clean across all 120 fits, with zero divergent transitions, no treedepth saturation, and E-BFMI greater than 0.3 for all chains.

\subsection{Per-person GAMM exploration}

We first apply the Stage 1 hierarchical GAMM to all three items. The factor-smooth interaction model is fit using {\tt bam} with basis dimension $k = 8$ per smooth term. For each participant, the model produces an estimated person-specific lag-effect curve $\hat{f}(\dt) + \hat{b}_i(\dt)$ over lag distances 1 through 14.

Table~\ref{tab:gamm-summary} summarizes the range of person-specific lag-1 effects estimated by Stage 1.

\begin{table}[ht]
\centering
\small
\begin{tabular}{lrrrrr}
\toprule
\textbf{Item} & \textbf{Mean} & \textbf{SD} & \textbf{Min} & \textbf{Max} & \textbf{Range} \\
\midrule
down & 0.385 & 0.168 & 0.064 & 0.896 & 0.832 \\
worried & 0.363 & 0.164 & 0.077 & 0.786 & 0.709 \\
energetic & 0.248 & 0.144 & 0.029 & 0.632 & 0.604 \\
\bottomrule
\end{tabular}
\caption{Person-specific lag-1 autoregressive effects from per-person factor-smooth GAMMs ($N = 40$ participants). For all three items, person-specific effects span an order of magnitude or more, demonstrating that between-person heterogeneity is substantial and not item-specific.}
\label{tab:gamm-summary}
\end{table}

For all three items, person-specific lag-1 effects span an order of magnitude or more, ranging from near zero to above 0.8. The between-person standard deviation is approximately 0.14 to 0.17 across items. These results establish that group-level estimation of a single lag-1 effect would mask substantial between-person variation in the actual dynamics, motivating the hierarchical Stage 2 model.

\begin{figure}[ht]
\centering
\begin{subfigure}{0.49\textwidth}
\centering
\includegraphics[width=\linewidth]{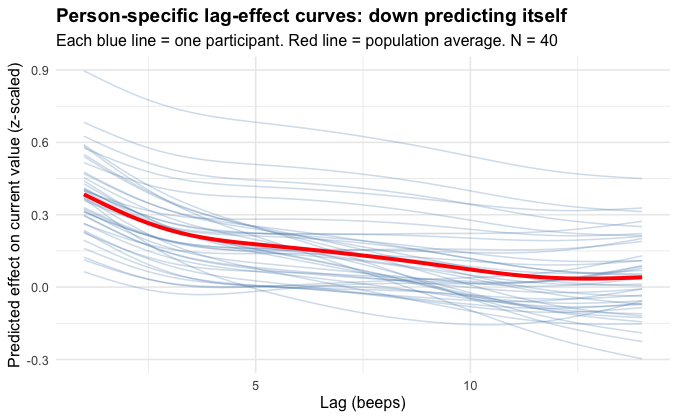}
\caption{Person-specific lag-effect curves (spaghetti plot). Each blue line is one participant's estimated profile; the red line is the population average.}
\label{fig:gamm-spaghetti}
\end{subfigure}
\hfill
\begin{subfigure}{0.49\textwidth}
\centering
\includegraphics[width=\linewidth]{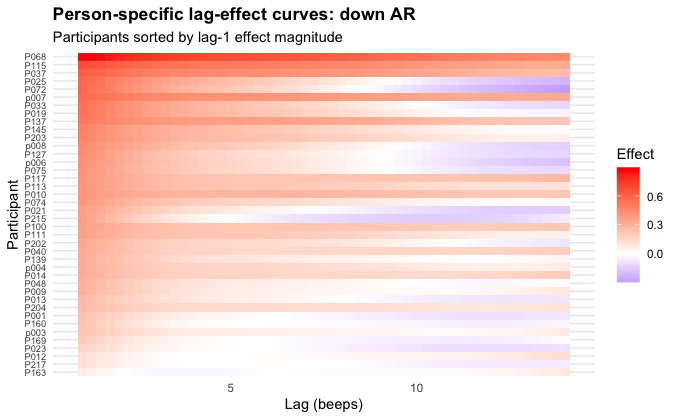}
\caption{The same per-person lag-effect estimates as a heatmap, with participants sorted top-to-bottom by lag-1 effect magnitude.}
\label{fig:gamm-heatmap}
\end{subfigure}
\caption{Person-specific lag-effect curves for the \textit{down} item, estimated by the Stage 1 hierarchical GAMM with factor-smooth interaction across the 40 participants. Panel (a) shows the heterogeneity unsorted; panel (b) sorts participants by lag-1 magnitude to reveal the gradient structure of person-specific decay profiles.}
\label{fig:stage1-gamm}
\end{figure}

Figure~\ref{fig:stage1-gamm} displays the resulting person-specific lag-effect curves for the {\it down} item in two complementary views: an unsorted spaghetti plot (panel a) and a sorted heatmap (panel b). The spaghetti view emphasizes the spread of person-specific curves around the population average; the heatmap view, with participants ordered top-to-bottom by lag-1 effect magnitude, reveals the gradient structure of person-specific decay profiles across the sample.

\subsection{Stage 2 hierarchical Bayesian estimation}

We next fit the lag-1 DTVEM-RE Stage 2 model to each of the three items. Table~\ref{tab:stan-summary} summarizes the posterior estimates.

\begin{table}[ht]
\centering
\small
\begin{tabular}{lrrrr}
\toprule
\textbf{Item} & $\hat\mu_a$ \textbf{[90\% CI]} & $\hat\tau_a$ \textbf{[90\% CI]} & $\hat\sigma_{\text{mean}}$ & \textbf{Stan-GAMM $r$} \\
\midrule
down      & 0.394 [0.346, 0.443] & 0.157 [0.122, 0.199] & 0.895 & 0.916 \\
worried   & 0.373 [0.322, 0.423] & 0.160 [0.125, 0.199] & 0.905 & 0.918 \\
energetic & 0.259 [0.219, 0.301] & 0.124 [0.090, 0.163] & 0.953 & 0.868 \\
\bottomrule
\end{tabular}
\caption{DTVEM-RE Stage 2 posterior summaries across three EMA items. All credible intervals for $\tau_a$ exclude zero, confirming that person-specific variation in lag-1 dynamics is statistically robust. Person-specific posterior means agree strongly with independent per-person GAMM estimates ($r = 0.87$ to $0.92$).}
\label{tab:stan-summary}
\end{table}

Three findings deserve emphasis. First, all 90 percent credible intervals for $\tau_a$ are bounded above zero across all three items, with point estimates of 0.12 to 0.16. This provides direct statistical evidence that between-person variation in lag-1 dynamics is robust and not noise. Second, the Stan posterior mean estimates of person-specific lag-1 coefficients agree strongly with the independent per-person GAMM estimates from Stage 1 ($r = 0.87$ to $0.92$). This cross-method validation rules out the possibility that the heterogeneity finding is an artifact of a specific statistical approach.

\begin{figure}[ht]
\centering
\includegraphics[width=0.95\textwidth]{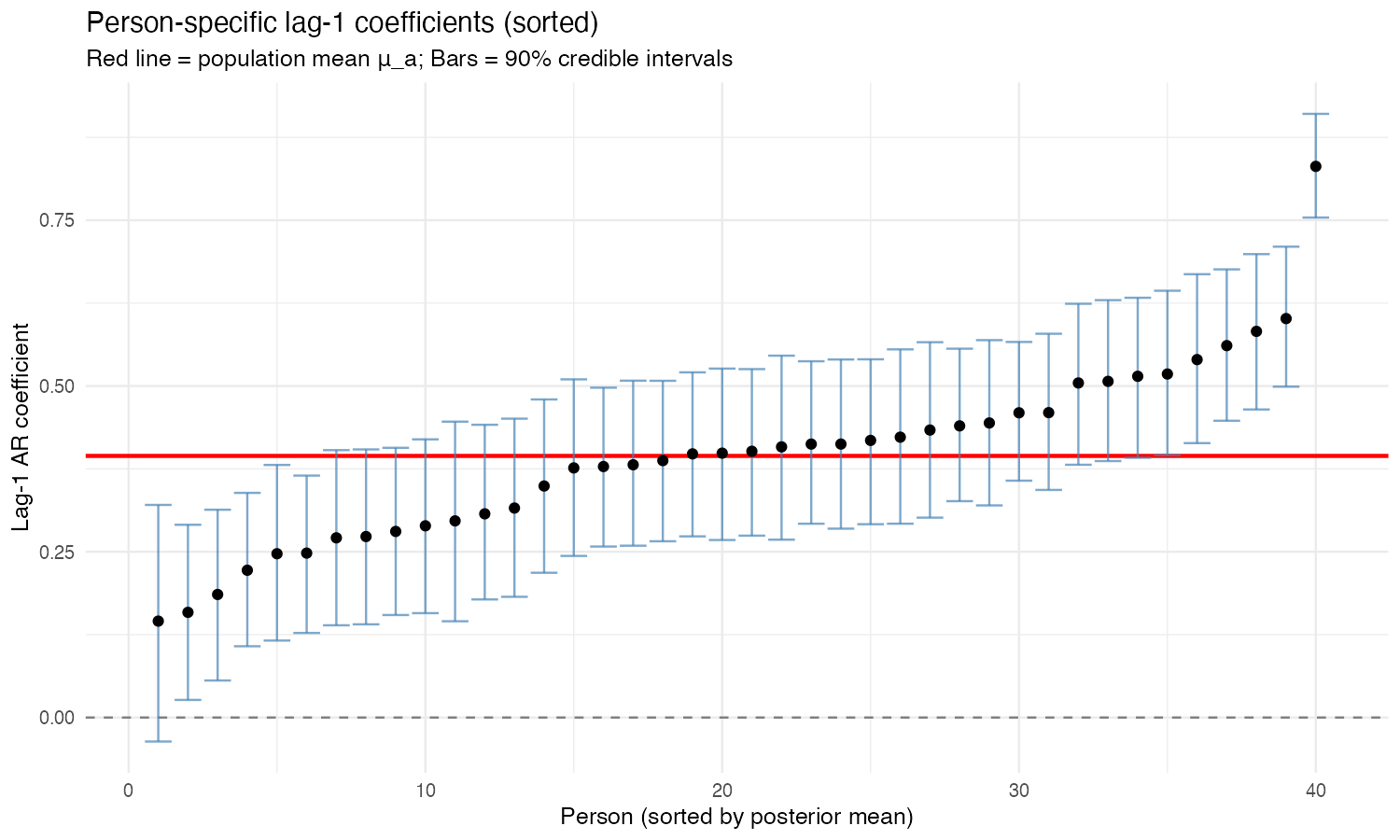}
\caption{Person-specific lag-1 autoregressive posteriors from the Stage 2 hierarchical Bayesian fit on the \textit{down} item. Each point is the posterior mean for one of the 40 participants; vertical bars show 90\% credible intervals. Participants are sorted left-to-right by posterior mean. The horizontal red line marks the population mean $\hat\mu_a = 0.394$; the dashed line at zero marks the no-dependence reference. Posterior means range from approximately $0.15$ to $0.83$, and 39 of 40 credible intervals exclude zero, indicating substantive person-specific autoregressive dependence across the sample.}
\label{fig:caterpillar}
\end{figure}

Third, the relative ordering of items by mean persistence is preserved: {\it down} and {\it worried} (both negative affect) show higher mean autoregressive coefficients than {\it energetic} (positive affect). We return to this pattern in the Discussion. Figure~\ref{fig:caterpillar} displays the person-specific lag-1 posterior distributions for the \textit{down} item, sorted by posterior mean. The figure also documents that 39 of 40 participants have 90 percent credible intervals on their lag-1 coefficient excluding zero, indicating that the autoregressive dependence is reliably non-zero for the substantial majority of individuals in the sample.

MCMC diagnostics for all three lag-1 fits were clean. The model with non-centered parameterizations on both $a_i$ and $\sigma_i$ produced zero divergent transitions, zero treedepth saturations, and minimum E-BFMI greater than 0.70 across all 12 chains (4 chains $\times$ 3 items). Total computation time was approximately 15 seconds per item.

\subsection{Held-out prediction comparison}

Following the protocol described in Section 3.9, we held out the last 12 beeps per participant on the {\it down} item and compared one-step-ahead predictions from four methods: group-only DTVEM, naive per-person DTVEM, a hierarchical simple (mlVAR-style) baseline, and DTVEM-RE in its full form. Performance is summarized in Table~\ref{tab:prediction}.

\begin{table}[ht]
\centering
\small
\begin{tabular}{lrrr}
\toprule
\textbf{Method} & \textbf{Mean log-lik} & \textbf{Total log-lik} & \textbf{RMSE} \\
\midrule
1. Group-only DTVEM       & $-$1.244 & $-$597.26 & 0.832 \\
2. Naive per-person       & $-$1.235 & $-$592.83 & 0.830 \\
3. Hierarchical simple    & $-$1.238 & $-$594.25 & 0.827 \\
4. DTVEM-RE (full)        & \textbf{$-$1.233} & \textbf{$-$591.81} & \textbf{0.827} \\
\bottomrule
\end{tabular}
\caption{Held-out predictive log-likelihood for the {\it down} item across four methods (480 held-out beeps from 40 participants). DTVEM-RE achieves the highest mean and total log-likelihood and the lowest root-mean-square error.}
\label{tab:prediction}
\end{table}

DTVEM-RE attains the highest log-likelihood and lowest RMSE across the four methods. In per-participant decomposition, DTVEM-RE outperforms naive per-person OLS in 25 of 40 participants (62.5 percent). However, the absolute magnitude of the advantage is modest, with differences across methods of approximately 0.01 log-likelihood units per beep. This modest advantage is consistent with the underlying structure: at $T \approx 110$ training beeps per person, naive per-person OLS is sufficiently stable that hierarchical pooling delivers only incremental gains. The substantive value of DTVEM-RE in this regime lies in calibrated uncertainty quantification (proper credible intervals on person-specific $a_i$) and in the multi-lag extension reported next, not in dramatic predictive supremacy.

\subsection{Multi-lag extension: the differentiating result}

We next refit the DTVEM-RE model in its multi-lag form, simultaneously estimating person-specific coefficients at lags 1, 2, and 3 with separate hyperparameters $(\mu_k, \tau_k)$ at each lag. The model specification is given in Section 3.5; computation took approximately 2 minutes per item.

\begin{table}[ht]
\centering
\small
\begin{tabular}{lccc|ccc}
\toprule
& \multicolumn{3}{c|}{\textbf{Population means ($\hat\mu_k$)}} & \multicolumn{3}{c}{\textbf{Between-person SDs ($\hat\tau_k$)}} \\
\textbf{Item} & lag 1 & lag 2 & lag 3 & lag 1 & lag 2 & lag 3 \\
\midrule
down      & 0.315 & 0.107 & 0.084 & \textbf{0.111}$^\star$ & \textbf{0.096}$^\star$ & \textbf{0.079}$^\star$ \\
worried   & 0.281 & 0.141 & 0.096 & \textbf{0.105}$^\star$ & \textbf{0.076}$^\star$ & \textbf{0.108}$^\star$ \\
energetic & 0.211 & 0.104 & 0.061 & \textbf{0.078}$^\star$ & \textbf{0.118}$^\star$ & \textbf{0.057}$^\star$ \\
\bottomrule
\end{tabular}
\caption{Multi-lag DTVEM-RE posterior means at lags 1, 2, and 3 for three EMA items. Starred ($^\star$) values for $\hat\tau_k$ have 90 percent credible intervals excluding zero. All nine $\hat\tau_k$ estimates exclude zero, establishing that person-specific variation in autoregressive dynamics is statistically robust at every lag examined.}
\label{tab:multilag-summary}
\end{table}

\begin{figure}[ht]
\centering
\begin{subfigure}{0.49\textwidth}
\centering
\includegraphics[width=\linewidth]{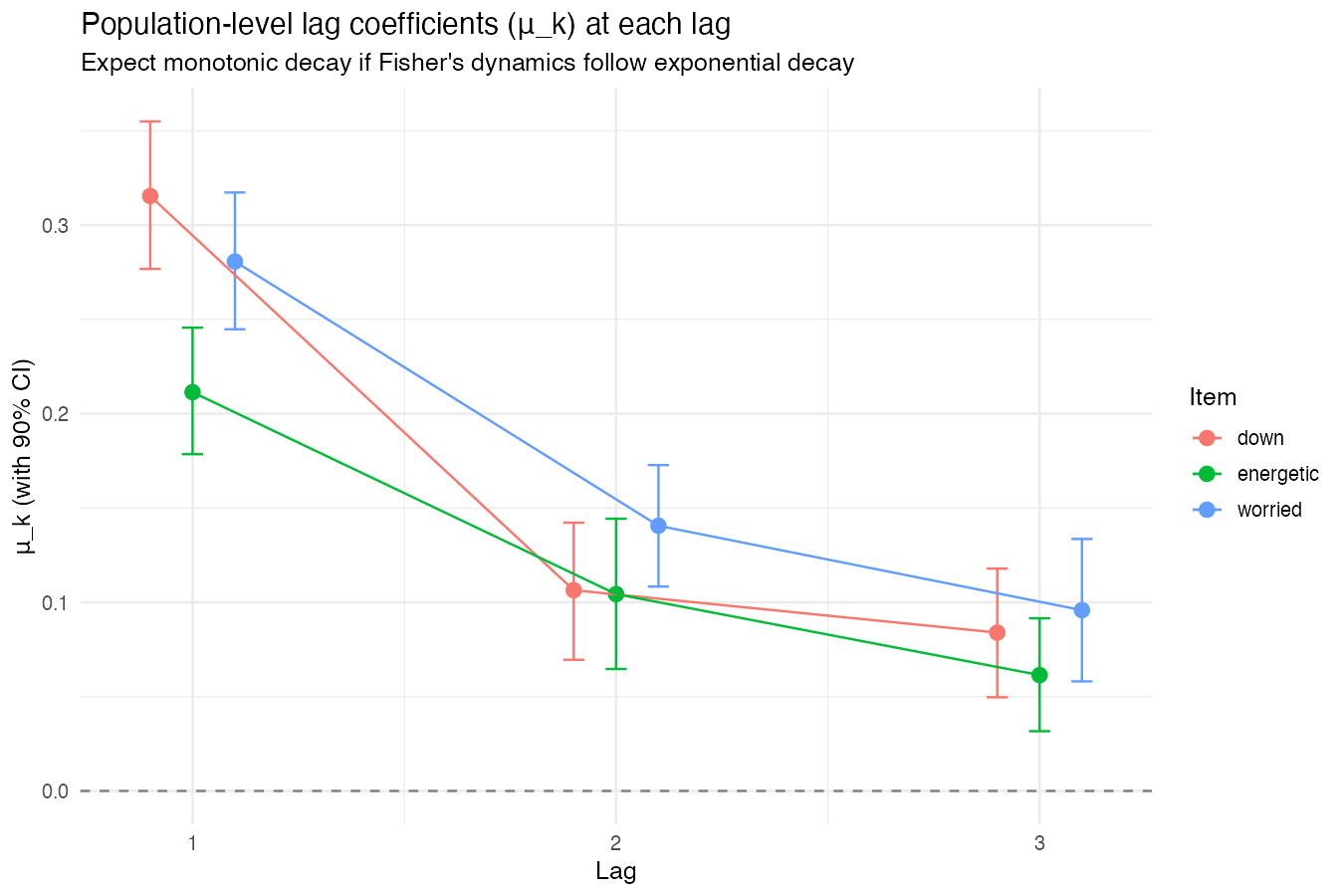}
\caption{Population-level mean coefficients $\hat\mu_k$ at lags 1, 2, and 3 for each item. All three items show monotonic decay of mean autoregressive effect across lags.}
\label{fig:multilag-mu}
\end{subfigure}
\hfill
\begin{subfigure}{0.49\textwidth}
\centering
\includegraphics[width=\linewidth]{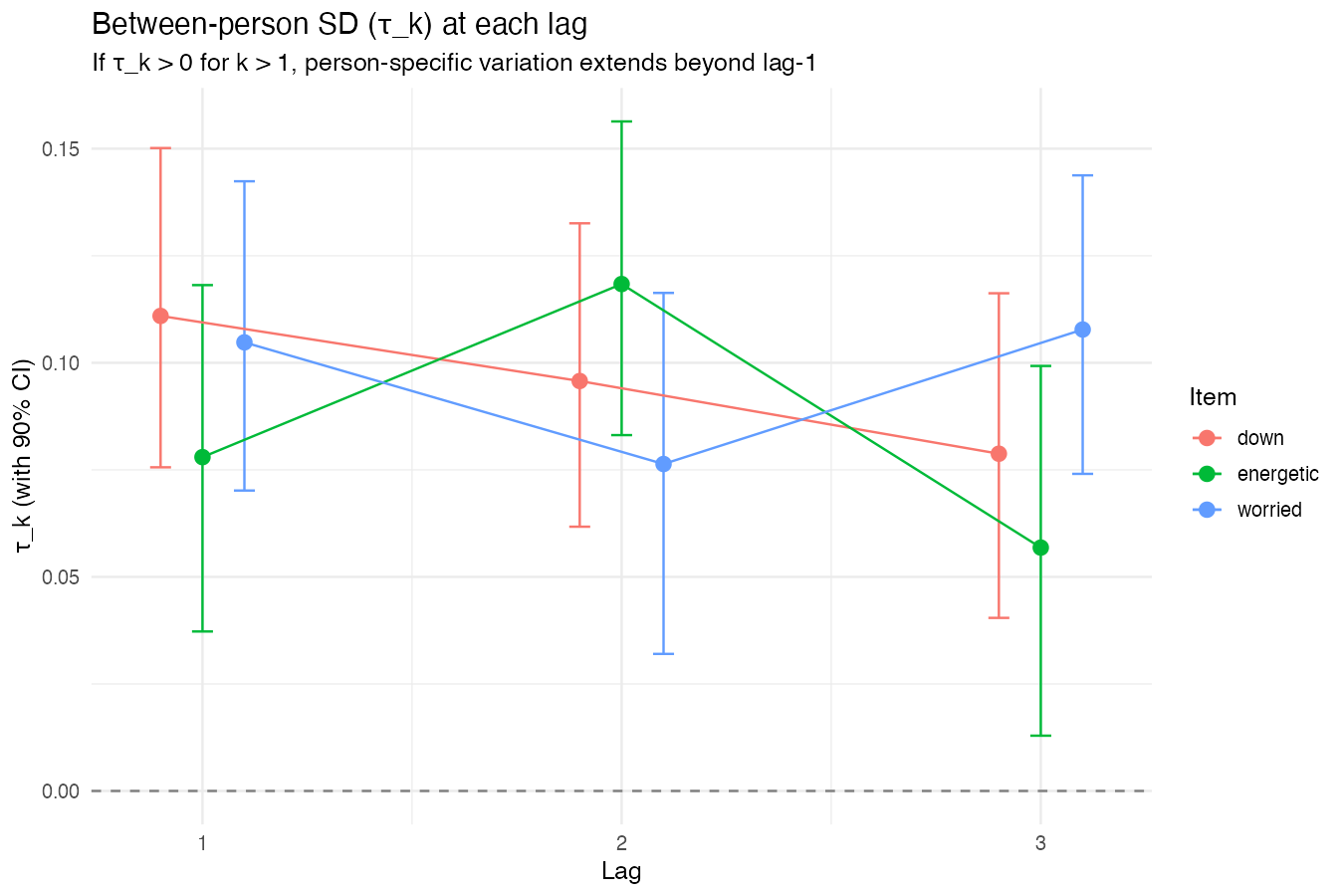}
\caption{Between-person SD $\hat\tau_k$ at lags 1, 2, and 3 for each item. All nine credible intervals exclude zero. The lag at which heterogeneity is largest differs across items: lag 1 for \textit{down}, lag 2 for \textit{energetic}, and lag 3 for \textit{worried}.}
\label{fig:multilag-tau}
\end{subfigure}
\caption{Multi-lag DTVEM-RE posterior summaries for the three EMA items. Points show posterior means; bars show 90\% credible intervals. The contrast between the two panels is the central methodological result of this paper: population-level means follow monotonic decay across lags (a), but between-person heterogeneity does not. All nine $\hat\tau_k$ credible intervals exclude zero, and the lag at which heterogeneity peaks differs by item---a structure that hierarchical methods placing random effects only on lag-1 coefficients are not designed to detect.}
\label{fig:multilag-results}
\end{figure}

The key result is in the right half of Table~\ref{tab:multilag-summary} and visualized in Figure~\ref{fig:multilag-results}. All nine $\hat\tau_k$ values, across three items and three lags, have 90 percent credible intervals that exclude zero. Person-specific variation in lag coefficients is statistically robust at every lag tested, not only at lag-1.

Furthermore, the lag at which heterogeneity is largest differs across items. For {\it down}, $\hat\tau_1 = 0.111$ is the largest, with heterogeneity monotonically declining at higher lags. For {\it worried}, $\hat\tau_3 = 0.108$ is the largest, with heterogeneity dipping at lag 2 and rising at lag 3. For {\it energetic}, $\hat\tau_2 = 0.118$ exceeds $\hat\tau_1 = 0.078$, indicating that individual differences in mood persistence are most pronounced at the two-beep horizon for positive affect.

This non-monotonic structure is outside the modeling scope of hierarchical VAR methods that place random effects only on lag-1 coefficients, such as mlVAR. A method that estimates a single $\tau_a$ at lag-1 would miss the larger heterogeneity at lag-2 for {\it energetic} or at lag-3 for {\it worried}. The shape of dynamic heterogeneity, not just its magnitude, varies across affect domains.

\begin{figure}[ht]
\centering
\includegraphics[width=0.85\textwidth]{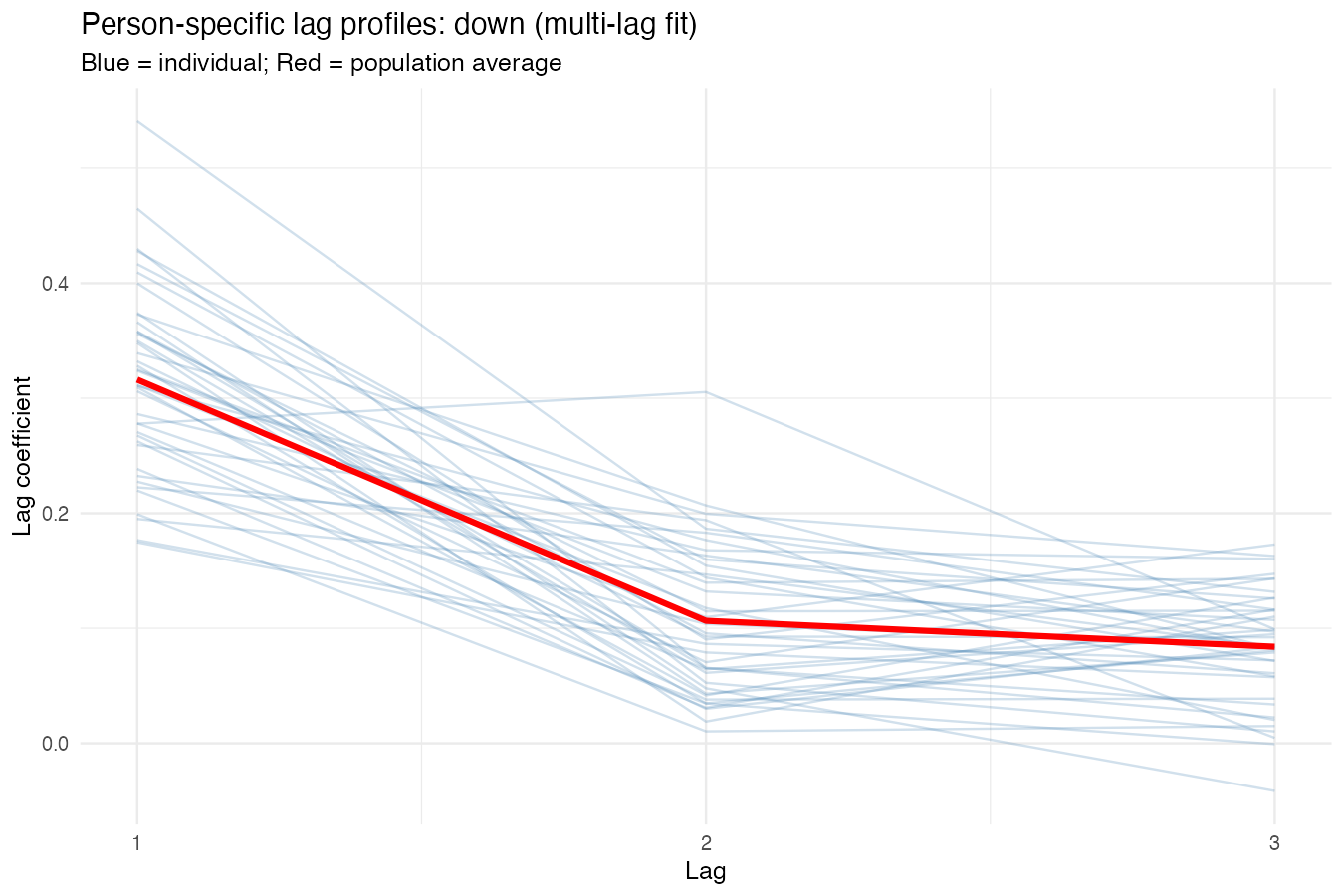}
\caption{Person-specific multi-lag profiles for the \textit{down} item from the multi-lag DTVEM-RE fit. Each blue line connects one participant's posterior mean lag coefficients at lags 1, 2, and 3; the red line shows the population means $(\hat\mu_1, \hat\mu_2, \hat\mu_3)$. The spread of person-specific profiles around the population mean at each lag illustrates the between-person heterogeneity quantified by the $\hat\tau_k$ estimates in Figure~\ref{fig:multilag-tau}. While most individuals show monotonic decay across lags, the rate and magnitude of decay vary substantially across the sample.}
\label{fig:multilag-profiles}
\end{figure}

Population-level means follow monotonic decay across lags (left half of Table~\ref{tab:multilag-summary}), consistent with the GAMM Stage 1 results. The aggregate dynamics are well-described by exponential decay at the group level; the novelty of DTVEM-RE is that it reveals heterogeneity in how individuals depart from that aggregate decay. Figure~\ref{fig:multilag-profiles} displays the person-specific multi-lag profiles for the \textit{down} item, with each blue line tracing one participant's three posterior means and the red line showing the population trajectory.

A natural follow-up question is whether person-specific lag coefficients are correlated across lags. We computed the Pearson correlations across the 40 participants between the posterior means at each lag pair. For \textit{down}: $r_{12} = 0.11$, $r_{13} = 0.06$, $r_{23} = 0.31$. For \textit{worried}: $r_{12} = 0.13$, $r_{13} = 0.14$, $r_{23} = 0.23$. For \textit{energetic}: $r_{12} = 0.45$, $r_{13} = 0.10$, $r_{23} = 0.28$. For the two negative-affect items, person-specific lag profiles are largely independent across lag distance: knowing a person's lag-1 dependence on \textit{down} or \textit{worried} carries little information about their lag-2 or lag-3 dependence. The positive-affect item shows a different pattern: \textit{energetic} exhibits a substantial lag-1/lag-2 correlation ($r = 0.45$), indicating that individuals with strong short-horizon dependence on positive activation also tend to be persistent at the next-shortest horizon. The contrast between the affect domains is itself a substantive observation enabled by the multi-lag specification and worth following up in larger samples.

\section{Discussion}

\subsection{Summary of contributions}

This paper introduced DTVEM-RE, a hierarchical random-effects extension of the Differential Time-Varying Effect Model that estimates person-specific multi-lag coefficients with shrinkage toward group-level means. We provided three sets of results. Simulation studies demonstrated clean recovery of the between-person variance parameter $\tau_a$ with absolute bias below 0.01 and near-nominal credible interval coverage at the sample sizes typical of EMA studies. Empirical demonstration on the Fisher (2017) dataset showed that person-specific lag-1 effects vary by an order of magnitude across affect items, that hierarchical Bayesian and independent GAMM estimates agree closely, and that DTVEM-RE attains the best one-step-ahead predictive performance among four hierarchical and non-hierarchical baselines, though by modest margins at the sample sizes available. Most importantly, the multi-lag extension showed that all nine between-person variance estimates across three items and three lags have credible intervals excluding zero, with the lag at which heterogeneity is largest differing across items in a manner that hierarchical methods placing random effects only on lag-1 are not designed to detect.

\subsection{Methodological implications}

The principal methodological implication is that the standard practice of estimating single group-level lag structures in intensive longitudinal data discards substantial between-person variation. For mood and anxiety dynamics in particular, the heterogeneity is large in magnitude (individual lag-1 effects ranging from near-zero to above 0.85), reliable in statistical terms (credible intervals excluding zero), and structurally complex (with non-monotonic distribution across lags).

DTVEM-RE provides a principled framework for engaging with this heterogeneity. The hierarchical priors enable stable estimation of person-specific coefficients even at the sample sizes typical of EMA studies ($T \approx 100$ to 150 per person). The cross-method agreement between Stan and independent GAMM estimates ($r = 0.87$ to $0.92$) provides reassurance that the recovered heterogeneity reflects genuine signal rather than methodological artifact.

\subsection{Clinical implications}

The present results are exploratory analyses of a single dataset with $N = 40$ participants and three affect items, and they do not license direct clinical recommendations. We outline below what the heterogeneity reported here \textit{could} mean clinically if it were to replicate in larger and more diverse samples, with the conditional emphasis throughout.

The substantive finding most relevant to clinical questions is that the lag at which an individual's affect shows its strongest self-prediction varies meaningfully across people. If this pattern replicates, it implies that the natural time scale of affect dynamics is itself a person-level feature rather than a universal property of the construct being measured. Two questions that downstream research could pursue in light of this. First, whether person-specific lag profiles are stable enough across measurement occasions to function as individual-difference variables: a person's lag profile is informative for any intervention or assessment decision only if it is stable enough to act on. Second, whether person-specific lag profiles relate to clinically meaningful variables --- symptom severity, treatment response, diagnostic boundary --- in ways that aggregate lag estimates obscure.

We emphasize that DTVEM-RE in its present form is a methodological tool for describing person-specific lag structure; the question of whether that description has clinical leverage is a separate empirical question and one this paper does not attempt to answer. Validation in independent samples, assessment of test-retest reliability, and prospective tests of intervention timing all lie beyond the present scope. We frame the multi-lag heterogeneity finding as a candidate phenomenon worth following up rather than as evidence for any specific clinical practice change.

\subsection{Substantive observations}

Two substantive patterns emerge from the empirical application that may warrant follow-up in future work, though we note that the current sample size ($N = 40$) is small for generalization. First, the two negative affect items in this analysis ({\it down} and {\it worried}) show numerically higher mean persistence than the positive affect item ({\it energetic}). Whether this reflects a general asymmetry between negative and positive affect persistence in clinical samples cannot be established from three items in one sample, but the pattern is consistent enough with prior affect dynamics literature to suggest it as a hypothesis worth testing in larger datasets. Second, the lag at which heterogeneity is largest differs across affect domains: depression-related dynamics show maximum heterogeneity at the shortest lag, positive affect at an intermediate lag, and anxiety-related dynamics at longer lags. The substantive interpretability of these patterns is limited by the small number of items examined, but the observation that they differ at all is methodologically relevant: any analysis that assumes heterogeneity is concentrated at a single lag will produce systematically distorted descriptions of those items for which that assumption is false.

\subsection{Limitations and future work}

Several limitations bear acknowledgment. First, the discrete-time assumption inherent in DTVEM-RE treats beeps as approximately equally spaced. Fisher's data have substantial within-day spacing variability (median 4.3 hours, with overnight gaps of 10 to 12 hours), which our analysis approximates by sequential beep numbering. Continuous-time methods such as ctsem \citep{driver2018ctsem} handle irregular spacing natively at the cost of imposing exponential decay. A direct comparison between DTVEM-RE and hierarchical ctsem on the same data would be informative and is a natural next step.

Second, the simulation study revealed a small finite-sample downward bias in the population mean $\mu_a$, traceable to the well-characterized small-sample bias in AR(1) estimation \citep{marriott1954bias, kendall1954bias}; the magnitude tracks the leading-order prediction at low-to-moderate heterogeneity and exceeds it at the highest heterogeneity condition examined (Section 4.1). This bias scales as $O(1/T)$ and would diminish in studies with longer per-person observation periods. For the heterogeneity-focused interpretation of DTVEM-RE results, the bias is uniform across persons within a sample and does not affect relative comparisons; researchers interested in absolute level interpretations should apply standard bias corrections.

Third, the held-out prediction advantage of DTVEM-RE over alternatives was modest at $T \approx 110$. Exploratory analysis at $T = 30$ showed that very small per-person samples can favor pooled estimation; we note this as a regime where prior choice matters and leave detailed investigation for future work. The regime where DTVEM-RE provides the largest predictive advantage relative to alternatives appears to be moderate per-person sample sizes (approximately $T = 50$ to 100), corresponding to typical week-long to month-long EMA designs.

Fourth, this paper has not yet incorporated diagnostic group information from the Fisher sample. The dataset's three subgroups (GAD-only, MDD-only, and comorbid) provide a natural test of whether person-specific lag profiles distinguish clinical presentations. We plan to add this analysis in subsequent work, conditional on hand-coding diagnostic labels from the original publication's Table 1.

\subsection{Conclusion}

DTVEM has become a standard tool for lag detection in intensive longitudinal data, but its group-level estimation assumption is at odds with the idiographic premise of much of the psychopathology literature that uses it. DTVEM-RE addresses this gap by extending both stages of DTVEM with hierarchical random-effects estimation, while retaining standard DTVEM as a fixed-effects special case. Empirical application demonstrates that person-specific variation in lag-1 effects is substantial in benchmark clinical data, and that multi-lag person-specific heterogeneity is statistically robust at every lag tested. This combination of nonparametric multi-lag exploration with hierarchical pooling on person-specific lag coefficients is, to our knowledge, methodologically new, and directly implements the random-effects extension named as future work by the original DTVEM authors.

\vspace{1em}

\section*{Code and data availability}

All R and Stan code for DTVEM-RE is available at \url{https://github.com/amartyacodes/DTVEM-RE}. The Fisher (2017) dataset is publicly available on the Open Science Framework at \url{https://osf.io/zefbc/}.

\vspace{1em}

\bibliographystyle{apalike}

\end{document}